\documentclass[conference]{IEEEtran}
\usepackage{times}
\pdfoutput=1
\let\chapter\section
\usepackage[numbers]{natbib}
\usepackage{multicol}
\usepackage[bookmarks=true]{hyperref}

\usepackage[linesnumbered,ruled]{algorithm2e}
\def\BState{\State\hskip-\ALG@thistlm}
\usepackage{pifont}
\usepackage{booktabs}

\usepackage{multirow}
\usepackage{graphicx}
\usepackage{amsfonts}
\usepackage{amsmath,amssymb,mathrsfs}
\usepackage{array}
\usepackage{flafter}
\usepackage{subfig}
\usepackage{verbatim}
\usepackage{color}
\usepackage{psfrag}
\usepackage{umoline}
\usepackage{hyperref}
\usepackage{indentfirst}

\newtheorem{problem}{Problem}

\newtheorem{lemma}{Lemma}

\allowdisplaybreaks[4]

\begin{document}

\title{Non-Gaussian SLAP: Simultaneous Localization and Planning Under Non-Gaussian Uncertainty in Static and Dynamic Environments}




%
\author{\authorblockN{Mohammadhussein Rafieisakhaei\authorrefmark{1},
Suman Chakravorty\authorrefmark{2} and
P.R. Kumar\authorrefmark{3}}
\authorblockA{\authorrefmark{1},\authorrefmark{3}Department of Electrical and Computer Engineering, \authorrefmark{2}Department of Aerospace Engineering
\\Texas A\&M University,
College Station, Texas 77840}
\authorblockA{\authorrefmark{1}Email: mrafieis@tamu.edu}}

\maketitle

\begin{abstract}
Simultaneous Localization and Planning (SLAP) under process and measurement uncertainties is a challenge. It involves solving a stochastic control problem modeled as a Partially Observed Markov Decision Process (POMDP) in a general framework. For a convex environment, we propose an optimization-based open-loop optimal control problem coupled with receding horizon control strategy to plan for high quality trajectories along which the uncertainty of the state localization is reduced while the system reaches to a goal state with minimum control effort. In a static environment with non-convex state constraints, the optimization is modified by defining barrier functions to obtain collision-free paths while maintaining the previous goals. By initializing the optimization with trajectories in different homotopy classes and comparing the resultant costs, we improve the quality of the solution in the presence of action and measurement uncertainties. In dynamic environments with time-varying constraints such as moving obstacles or banned areas, the approach is extended to find collision-free trajectories. In this paper, the underlying spaces are continuous, and beliefs are non-Gaussian. Without obstacles, the optimization is a globally convex problem, while in the presence of obstacles it becomes locally convex. We demonstrate the performance of the method on different scenarios.
\end{abstract}

\IEEEpeerreviewmaketitle
\vspace{-3pt}
\section{Introduction}\label{sec:intro}

Uncertainty is a feature of many robotic systems. Whether it is navigation underwater, aerially, or in a cluttered environment, uncertainty presents itself as an inevitable part of the problem. Dealing with uncertainty in any situation is a challenge. For a robust and reliable plan, the controller must solve a stochastic control problem \cite{Kumar-book-86,kumar2014control}, which can be formulated as a Partially Observed Markov Decision Process (POMDP) \cite{Astrom65}, \cite{Smallwood73}, \cite{Kaelbling98}. However, unless the domain of the planning problem is finite and small, finding the optimal control law in a general set-up is extremely difficult \cite{Papadimitriou87,Madani99}. This is because the planner needs to compute an offline policy function for all conditional probability distributions of the state, referred to as belief states. Approaches that solve the problem by either building a decision tree of discretized observation-action pairs \cite{GShaniJPineau13,Pineau03}, or maintaining a global discretization of the value function over continuous spaces, suffer from the exponential complexity known as the curse of dimensionality \cite{Kaelbling98,Zhou01}. A more significant problem is the exponential growth of the number of trajectories corresponding to action-observation pairs, known as the curse of history \cite{Pineau03}.

In contrast, deterministic motion-planning methods assume no uncertainty and aim to obtain collision-free paths. In these methods, the dynamics of the system and any kinds of uncertainty involved in the problem have no impact on the final trajectory. Falling in this category are sampling-based approaches, such as Rapidly exploring Random Tree (RRT) \cite{Lavalle01-RRT, Karaman.Frazzoli:IJRR11} and Probabilistic RoadMap (PRM)-based \cite{Kavraki96,Kavraki98} methods; and trajectory optimization-based algorithms, such as Covariant Hamiltonian Optimization for Motion Planning (CHOMP) \cite{zucker2013chomp} and TrajOpt \cite{schulman2014motion} which aim to obtain a collision-free trajectory. 

However, real world systems contain both motion and sensing uncertainty and require belief space planning. Assuming the belief is fully parameterized by a mean and covariance is a well-entrenched method in the literature. Early extensions of Linear Quadratic Gaussian (LQG)-based methods \cite{bertsekas1995dynamic}, such as iterative LQG (iLQG) \cite{todorov2005generalized}, restricted attention to process uncertainty by assuming full observations. By incorporating observation uncertainty, roadmap-based \cite{Prentice09, Huynh09} and tree-based methods \cite{van2011lqg} apply LQG methodology to find a locally optimal nominal path, rather than constructing a trajectory. \citet{Berg11-isrr} use stochastic Differential Dynamic Programming (sDDP) to extend LQG-MP \cite{van2011lqg} to roadmaps, which is then extended in \cite{van2012motion} utilizing iLQG to perform the belief value-iterations. Methods described in \cite{Toit10,Platt10ML} perform filtering during the planning stage using the most likely observation in order to make the belief propagation deterministic. Algorithms that consider Monte-Carlo simulations on future observations, or methods such as \cite{agha2011firm, Berg11-isrr} which consider all future observations, can perform better. However, LQG-based methods always require an initial feasible solution, which can limit trajectories and affect planning. In Feedback-Based Information RoadMap (FIRM) \cite{agha2013firm}, a belief-MDP problem is solved over a belief-graph. This method is advantageous because it breaks the curse of history; however, its belief stabilizations are LQG-based. In LQG-based methods, other than the Gaussianity assumption of the uncertainty (which is insufficient in many situations), computational cost and scalability is a concern. In case of large deviations, unpredicted error accumulations force the controller to re-plan. Hence, methods that provide an output trajectory instead of a feedback policy must have a fast re-planning algorithm. Thus, application of methods, such as \cite{Berg11-isrr} with high computational complexity becomes limited to problems of small domains. 

Receding Horizon Control (RHC) or Model Predictive Control (MPC)-based methods \cite{garcia1989model, mayne2014model} are closely related methods, where the optimization problem involved in the inner control loop is required to compute quickly. In these methods, at each time-step, a finite-horizon optimal control problem results in an optimal open-loop sequence of controls for the current state of the system whose first element is applied to the plant \cite{mayne2000constrained, mayne1990receding}. An appealing advantage of MPCs is that, unlike POMDP solvers that obtain an offline feedback policy prescribing the control for all (belief) states, MPCs have a natural online planning procedure for the current state of the plant. Particularly, in problems with changing environments with moving objects, offline plans can become unreliable after the execution of a few steps of planned actions, thus exploring only the relevant portion of the belief space. Deterministic MPCs have received the most attention in the literature, and stochastic MPCs are still under development. A large body of the literature focuses its attention on robust-MPCs and optimization in a tube of trajectories generated by propagating the initial state with several samples of process uncertainties \cite{jiang2001input, limon2009input}. These methods can lead to extreme conservativeness \cite{Lazar2013TAC}. Another class of methods tackles the problem with linear process and observation models. However, these methods can lead to either non-convex or high-computational costs \cite{hokayem2010stable}. In problems with process uncertainty, Monte-Carlo based methods \cite{bernardini2009scenario, kantas2009sequential}, and related methods such as the scenario approach \cite{calafiore2006scenario}, have also been successful in providing probabilistic guarantees with high confidence for convex problems. However, in the presence of obstacles, most problems become inherently non-convex which limits the application of such methods in robotics problems.

In this paper, a stochastic MPC is proposed for planning in the belief space. We define a finite-horizon optimal control problem with a terminal constraint that samples from an initial non-Gaussian belief and maps belief state samples to observation samples through usage of the observation model. The actions are planned with the aim of compression of the ensemble of observation trajectories. Therefore, the filtering equation, which usually poses a heavy burden on stochastic control problems, is avoided during the planning stage. Additionally, the ``most likely observation'' assumption is not used. The core problem in a convex environment is convexified for common non-linear observation models. Moreover, non-convex constraints are respected softly with a combination of line integration of barrier functions along the state trajectory. In a static environment, we apply the proposed optimization over trajectories in different homotopy classes to find the lowest cost, smooth, collision-free trajectory with uncertainty reduction and minimum effort. It should be noted that, although the optimization is initialized with trajectories, our algorithm constructs a new trajectory, since the underlying trajectory only defines the homotopy class of the trajectory and is morphed continuously into a locally optimal one. Moreover, as elaborated in the simulations, the initial trajectory need not be completely feasible; if it is feasible, the optimization avoids morphing towards infeasible regions. In dynamic environments where the obstacles can move or banned areas can change over time, the optimization problem is modified to respect the time-varying constraints via time-varying barrier functions. Therefore, in neither the static nor dynamic cases does the optimization vector size changes and the decision variables remain solely as the control variables. Our approach, which simplifies the planning stage and reduces the computational complexity for solving the open-loop optimal control problem, significantly improves online navigation under uncertainty when compared to previous methods. Planning for uncertainty and optimization becomes more efficient by reducing computational costs by utilizing our stochastic MPC, which then becomes inherently flexible and scalable; thus, enabling its usage in both static and dynamic environments. This flexibility allows our stochastic MPC to increase its scalability in the static environment, thereby moving from locally-optimal solutions towards a better approximation of a globally-optimal approach by applying the algorithm over multiple homotopy classes. Furthermore, once established, the scalability can make possible the treatment of dynamic environments with time-varying constraints or moving obstacles, which makes the algorithm suitable for online planning.


\section{Problem Definition}\label{sec:Problem}
In this section, we provide the general problem and other definitions.

\emph{System equations:} We denote the state of the system by $\mathbf{x}\in \mathbb{X} \subset\mathbb{R}^{n_x}$, control action by $\mathbf{u}\in \mathbb{U}\subset\mathbb{R}^{n_u}$, and observation vector by $\mathbf{z}\in \mathbb{Z}\subset\mathbb{R}^{n_z}$. We only consider holonomic systems in this paper, where the general system equations are $ \mathbf{x}_{t+1}=f(\mathbf{x}_{t},\mathbf{u}_{t},\boldsymbol{\omega}_{t}) $ with $ \mathbf{z}_{t}=h(\mathbf{x}_{t})+\boldsymbol{\nu}_{t} $, 
where $ \{\boldsymbol{\omega}_t\} $ and $ \{\boldsymbol{\nu}_t\} $ are zero mean independent, identically distributed (iid) mutually independent random sequences, $f:\mathbb{X}\times\mathbb{U}\times\mathbb{R}^{n_x}\rightarrow\mathbb{X}$ defines the process (motion) model, and $h:\mathbb{X}\rightarrow\mathbb{Z}$ specifies the observation (measurement) model.

\textit{Belief (information state):} The conditional distribution of $ \mathbf{x}_t $ given the data history up to time $ t $, called the belief $ b_t:\mathbb{X}\times\mathbb{Z}^{t}\times\mathbb{U}^{t}\rightarrow\mathbb{R} $. It is defined as $ b_{t}(\mathbf{x},\mathbf{z}_{0:t},\mathbf{u}_{0:t-1}, b_0):=p_{\mathbf{X}_t|\mathbf{Z}_{0:t};\mathbf{U}_{0:t-1}}(\mathbf{x}|\mathbf{z}_{0:t};\mathbf{u}_{0:t-1};b_0) $, and denoted by $ b_t(\mathbf{x}) $ or $ b_t $ in this paper \cite{Sondik71,Kumar-book-86,kaba2014bounds,kazempour2011utilization}. We use a non-Gaussian particle filter representation of a belief in belief space, $ \mathbb{B} $, by taking a number $ N $ of state samples $\{\mathbf{x}^i_t\}_{i=1}^{N}$ with importance weights $\{w^i_t\}_{i=1}^{N}$ \cite{Thrun2005,CrissanDoucet02, rafieisakhaei2016feedbackICRA,rafieisakhaeiline} and $ b_{t}(\mathbf{x})\!\!\approx\!\!\sum_{i=1}^{N}w^{i}_{t}\delta(\mathbf{x}-\mathbf{x}^{i}_t)$ where $\delta(\cdot)$ is the Dirac delta mass. It can be proven that by increasing the number of particles to infinity, the distribution of the particles tends to the true filtering distribution \cite{Doucet01Book}. In our planning problem, we employ for the Maximum-A-Posteriori (MAP) state estimate defined as $\hat{\mathbf{x}}_t= \arg\max_{\mathbf{x}\in \mathbb{X}}b_t(\mathbf{x}) $. The belief dynamics follow Bayesian update equations, summarized as a function $ \tau\!:\!\mathbb{B}\!\times\!\mathbb{U}\!\times\!\mathbb{Z}\rightarrow\mathbb{B} $, where $ b_{t+1}=\tau(b_{t},\mathbf{u}_t,\mathbf{z}_{t+1}) $ \cite{Bertsekas07}.

\begin{figure*}
  \centering
  {\includegraphics[width=\textwidth]{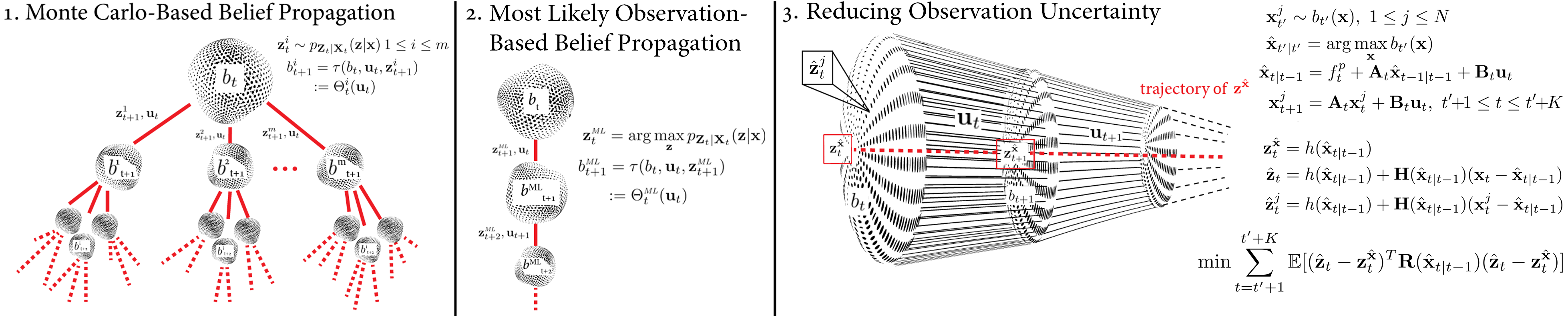}}
  \caption{Comparison of our method (3) with the traditional belief propagation methods (1 and 2).\label{fig:belief_propagarion} }\vspace{-10pt}
\end{figure*}

\begin{problem}\label{problem:General optimal control problem}
\textbf{\textit{General optimal control problem}}. For a holonomic system, given the initial belief state $ b_{t'} $ and a goal state $ \mathbf{x}_g $, we define the following optimal control problem:
\begin{subequations}\label{eq:main Problem}
\begin{align}
\nonumber \min_{\pi}~\mathbb{E}_{\pi}[&\sum_{t=0}^{K-1}c_t^{\pi}(\mathbf{x}_t,\mathbf{u}_t)+c_K(\mathbf{x}_K)]
\\ s.t.~b_{t+1}&=\tau(b_{t},\mathbf{u}_t,\mathbf{z}_{t+1})\label{eq:filtering}
\\\mathbf{x}_{t+1}&=f(\mathbf{x}_{t},\mathbf{u}_{t},\boldsymbol{\omega}_{t})
\\\mathbf{z}_{t}&=h(\mathbf{x}_{t})+\boldsymbol{\nu}_{t}
\\\phi_t^j({\mathbf{x}}_{t})&<0,~ j\!\in\![1,n_c],
\end{align}
\end{subequations}
where the optimization is over feasible policies, and $ \pi=\{\pi_{0}, \cdots, \pi_{t}\} $ where $ \pi_{t} :\mathbb{Z}^{t+1}\rightarrow \mathbb{U} $ specifies an action given the output, $ \mathbf{u}_{t}=\pi_{t}(\mathbf{z}_{0:t}) $. Moreover, $ c^{\pi}_t(\cdot,\cdot):\mathbb{X}\times\mathbb{U}\rightarrow\mathbb{R} $ is the one-step cost function, $ \phi_t^j({\mathbf{x}}_{t})<0 $ are $ n_c(K\!+\!1) $ inequality constraints, $ c_K^{\pi}(\cdot):\mathbb{X}\rightarrow\mathbb{R} $  denotes the terminal cost, and the expectation is taken with respect to policy $ \pi $.
\end{problem} 

Now that we have defined our problem, we present the proposed solution for different variations of this problem.

\section{Our Solution}\label{sec:Our Solution}

\subsection{Belief Space Planning}\label{subsec:Belief Space Planning}

Calculation of the solution for the belief space problem \ref{problem:General optimal control problem} in general is intractable. One attempt to make the problem more tractable is to restrict the policy to strategies such as RHC, reduce the closed-loop optimization to an open-loop optimal control problem, and close the feedback loop by re-planning after execution of each action to account for updated belief. Even though the open-loop problem does not provide a feedback policy for all beliefs, it provides an open-loop optimal control sequence that still involves optimization in the belief space, which is a function space. Particularly, since the observation space is continuous, the constraint of equation \eqref{eq:filtering} makes the problem computationally intractable. However, we will define our cost function such that the filtering equation is incorporated in the cost function. It can be shown that equation \eqref{eq:filtering} can be omitted to make the optimization more tractable, while the approximate solution maintains its quality.

\textit{Linearized system equations:} We linearize the process model around a nominal trajectory $ \{\mathbf{x}^{p}_{t}\}_{t=t'}^{t'+K}, \{ \mathbf{u}^{p}_{t}\}_{t=t'}^{t'+K-1} $ for a lookahead time horizon of $ K $ and linearize the observation model about the \textit{MAP state} trajectory, to obtain:
\begin{subequations}\label{eq:linearized system1}
\begin{align}
\mathbf{x}_{t+1}&=f^{p}_{t}+\mathbf{A}_t\mathbf{x}_t + \mathbf{B}_t\mathbf{u}_t +\mathbf{G}_t\boldsymbol{\omega}_t\label{eq:linearized process model}\\
\mathbf{z}_{t}&=h(\hat{\mathbf{x}}_{t})+\mathbf{H}(\hat{\mathbf{x}}_{t})(\mathbf{x}_t-\hat{\mathbf{x}}_{t})+\boldsymbol{\nu}_t,\label{eq:linear observation}
\end{align}
\end{subequations}
where $ f^{p}_{t}:= f(\mathbf{x}^{p}_{t},\mathbf{u}^{p}_t,0) -\mathbf{A}_t\mathbf{x}^{p}_{t} - \mathbf{B}_t\mathbf{u}^{p}_{t}$, and the matrices $ \mathbf{A}_t=\partial f(\mathbf{x},\mathbf{u},0)/\partial \mathbf{x}|_{ \mathbf{x}^{p}_{t}, \mathbf{u}^{p}_{t} } $, $ \mathbf{B}_t=\partial f(\mathbf{x},\mathbf{u},0)/\partial \mathbf{u}|_{ \mathbf{x}^{p}_{t}, \mathbf{u}^{p}_{t} } $, and $ \mathbf{G}_t=\partial f(\mathbf{x},\mathbf{u}_t,0)/\partial \boldsymbol{\omega}|_{ \mathbf{x}^{p}_{t}, \mathbf{u}^{p}_{t} } $ are constant during each time step (but can be time-dependent). In our planning problem we construct an optimal trajectory. Therefore, in holonomic systems and under saturation constraints, the initial nominal trajectory can be morphed to differ significantly from the final optimized trajectory. However, it is important that although we linearize the observation model, the resultant Jacobian $ \mathbf{H}(\hat{\mathbf{x}}_{t})=\partial h(\mathbf{x},0)/\partial \mathbf{x}|_{ \hat{\mathbf{x}}_{t} } $ is \textit{non-linear}, state-and-time dependent for many observation models. In other words, $ \mathbf{H}(\hat{\mathbf{x}}_{t}) $ is a function of the control variables.

\textit{Incremental cost} $ c(\cdot,\cdot):\mathbb{X}\times\mathbb{U}\rightarrow\mathbb{R} $ is defined as:
\begin{align}\label{eq:incorporate I}
\!\!\!c(\mathbf{x}_t,\mathbf{u}_t)\!\!=\!\!\mathbb{E}_{b_t}[\!(\mathbf{x}_t\!-\!\hat{\mathbf{x}}_{t})^T\mathbf{W}(\hat{\mathbf{x}}_{t})(\mathbf{x}_t\!-\!\hat{\mathbf{x}}_{t})\!+\!\mathbf{u}_{t-1}^T\mathbf{V}_t^u\mathbf{u}_{t-1}\!],
\end{align}
where $ \mathbf{W}(\hat{\mathbf{x}}_{t}):=\mathbf{H}(\hat{\mathbf{x}}_{t})^{T}\mathbf{R}(\hat{\mathbf{x}}_{t})\mathbf{H}(\hat{\mathbf{x}}_{t})\succ 0 $ and $ \mathbf{V}_t^{u}\succ 0 $ are positive definite matrices, and $ \mathbf{R}:\mathbb{X}\rightarrow\mathbb{R}^{n_z\times n_z} $ is a proper weighting matrix, to be defined later. Conceptually, the special definition of the $ \mathbf{W} $ matrix converts the weighted error of state particles to a weighted error of observation particles. Doing this, we are incorporating the uncertainty reduction in the cost, \textit{avoiding the filtering equation} as a constraint in the optimization, which breaks the computational cost of planning problem, avoids introduction of new decision variables, avoids propagation of the belief using most likely observations, and distinguishes our method from LQG-based methods and similar particle filter-based methods. This cost seeks to reduce the dispersion in the ensemble of the observation trajectories in terms of the weighted covariance. In other words, the minimization seeks to reduce the uncertainty in the predicted observation or equivalently in estimation, which translates itself to shrinking the support of belief distribution. 

\textit{Comparison of our method with traditional approaches:} Figure \ref{fig:belief_propagarion} graphically compares our method with traditional methods in the literature that tackle the \textit{open-loop} problem. In order to perform the filtering equation, a previous belief and action, and a current observation are required. In the planning stage, where the controller obtains the best sequence of future actions, a current belief is given; however, all that is known about the future observation is a likelihood distribution. As shown in this figure, in classic methods, the initial belief is propagated using finitely many samples of the observation obtained from the likelihood distribution. Therefore, a decision tree on the future predicted beliefs is constructed and so that the optimizer can obtain the best action for each height of the tree. Overall, the first method is computationally intensive. In the second popular class of methods, only most likely observation is utilized to perform the filtering equations and propagate the belief. This method can be less accurate than the latter, and although it provides a less expensive optimization, nevertheless, the filtering equation is part of the optimization constraints which limits the application of this method to small domains. However, in our method, once the samples of the initial belief are propagated via the predicted model of the system, they are converted into observation particles by a proper usage of the observation model. Thus, a rope-like bundle of propagated observation particle strands is constructed using the initial belief samples and with the advantage of a particular defined cost function, the dispersion in the strands is minimized. Hence, the optimization not only morphs the rope towards regions that provide observations, but also seeks to compress the bundle towards the end of the horizon. As a result of reduced uncertainty in observation bundle, the belief itself shrinks and the same results are obtained without performing the filtering equation. Therefore, using this idea, the main computational burden of the problem is broken and the much cheaper optimization yields the desired results. We provide more details as we proceed in the paper.
\vspace{-4pt}
\subsection{Convex Environment}\label{subsec:sol convex environment}
In this subsection, we provide our solution to the first variant of the problem \ref{problem:General optimal control problem}. Our assumption in this variant is that the inequalities $ \phi_t^j(\hat{\mathbf{x}}_{t})<0 $ can only be convex, such as the ones that define the boundary of the problem or saturation constraints on the control inputs.

\textit{Approximating the expectation in the cost:} Using a particle representation, we can simplify the cost function by taking the expectation over the state samples. We propagate the initial belief using a noiseless process model \eqref{eq:linearized process model} to obtain $ \mathbf{x}^{i}_{t_2}\!\!-\hat{\mathbf{x}}_{t_2}\!\!=\!\!\tilde{\mathbf{A}}_{t_1:t_2-1}(\mathbf{x}^{i}_{t_1}-\hat{\mathbf{x}}_{t_1}) $, where $\{\mathbf{x}^{i}_{t_1}\}^{k}_{i=1}$ are the set of resampled particles at time step $ t_1 $, $ \tilde{\mathbf{A}}_{t_1:t_2}: = \Pi_{t=t_1}^{t_2} \mathbf{A}_{t} $, defined for $ t_1\le t_2 $. Thus, the expectation in \eqref{eq:linearized process model} is approximated as:
\begin{align}\label{eq:for complexity}
\nonumber\sum\limits_{t=t'+1}^{t'+K}\!\!\![\frac{1}{N}&\sum_{i=1}^{N}[(\mathbf{x}^i_{t'}-\hat{\mathbf{x}}_{t'})^T\tilde{\mathbf{A}}_{t':t-1}^T\mathbf{W}(\hat{\mathbf{x}}_{t})\tilde{\mathbf{A}}_{t':t-1}(\mathbf{x}^i_{t'}-\hat{\mathbf{x}}_{t'})]\\&+\mathbf{u}_{t-1}^T\mathbf{V}_t^u\mathbf{u}_{t-1}].
\end{align}

Next, we utilize the $ \mathbf{R} $ matrix to convexify the cost function for non-linear observations.
\begin{lemma}\label{lemma 1}\textit{Scalar observation} Suppose $ \mathbf{d}=(d_1,\cdots,d_{n_x})^T\in\mathbb{R} $ and $ h(\mathbf{x}):\mathbb{X}\rightarrow\mathbb{R} $ differentiable. If $ l:\mathbb{X}\rightarrow\mathbb{R} $ defined as $ l(\mathbf{x}):=\sqrt{R(\mathbf{x})}(\mathbf{d}\cdot \mathbf{H}(\mathbf{x})^{T})
$, is convex or concave in $ \mathbf{x} $, then $ g:\mathbb{X}\rightarrow\mathbb{R}_{\ge 0} $, where $ g(\mathbf{x}):=\mathbf{d}^T\mathbf{H(x)}^TR(\mathbf{x})\mathbf{H(x)}\mathbf{d} $ is a convex function of $ \mathbf{x} $, where $ \mathbf{H}(\mathbf{x}):=\nabla h|_{\mathbf{x}} $ is the Jacobian of $ h $.

\end{lemma}
\noindent Lemma \ref{lemma 1} can be proved observing that $ g(\mathbf{x}) = (l(\mathbf{x}))^2 $. Moreover, $ \mathbf{R} $ can be deigned to be a positive and increasing function of the distance from the information sources other than the features in Lemma \ref{lemma 1}. This is mainly because, most of common observation models such as range and bearing are functions of such a distance. Therefore, more distant states from the information sources get more penalized, penalizing the corresponding observation trajectories, as well.

\textit{Convexifying the first term of the cost:} Given the differentiable observation model $ \mathbf{h}(\mathbf{x}) = \left[
{h}_j(\mathbf{x})
\right] $, and its Jacobian
$ \mathbf{H}(\mathbf{x})=\left[\begin{aligned}
{\mathbf{H}}_j(\mathbf{x})\end{aligned}\right], $
where $ \mathbf{H}_j(\mathbf{x}) = \left[\begin{aligned}
\frac{\partial {h}_j(\mathbf{x})}{\partial x_1}, \cdots, \frac{\partial h_j(\mathbf{x})}{\partial x_{n_x}}
\end{aligned}\right] $ for $ 1\le j\le n_z $, if $ \mathbf{R}(\mathbf{x})= \mathrm{diag}({R}_j(\mathbf{x}))$ is the diagonal matrix of $ {R}_i(\mathbf{x}) $'s corresponding to (uncorrelated) observations, extending the definition of $ g $ to include matrix $ \mathbf{R} $ we have $
\mathbf{d}^T\mathbf{H}(\mathbf{x})^{T}\mathbf{R}(\mathbf{x})\mathbf{H}(\mathbf{x})\mathbf{d}=\sum\limits_{j=1}^{n_z}R_j(\mathbf{x})(\mathbf{d}\cdot\mathbf{H}_{j}(\mathbf{x})^{T})^2,
$ which is a sum of positive convex functions as in Lemma \ref{lemma 1}. Thus, the first term of the cost in equation \eqref{eq:incorporate I} can be convexified. In our cost function, vector $ \mathbf{d} $ represents any of the vectors $ \mathbf{e}^{i}_{t} := \frac{1}{\sqrt{N}}\tilde{\mathbf{A}}_{t':t-1}(\mathbf{x}^i_{t'}-\hat{\mathbf{x}}_{t'})\in\mathbb{R}^{n_x} $ for $ 1\le i\le N $, summarized in vector $ \mathbf{e}_t := (\mathbf{e}^{1^{T}}_{t}, \mathbf{e}^{2^{T}}_{t}, \cdots, \mathbf{e}^{N^{T}}_{t})^T\in\mathbb{R}^{Nn_x} $ which is constant at each time step $ t $.

\begin{problem}\label{problem:Core convex problem in convex feasible space}
\textbf{\textit{Core convex problem in convex feasible space}}. Define $ \bar{\mathbf{W}}(\hat{\mathbf{x}}_{t}) := \mathrm{BlockDiag}(\mathbf{W}(\hat{\mathbf{x}}_{t})) $ with $ N $ equal diagonal blocks of $ {\mathbf{W}}(\hat{\mathbf{x}}_{t}) $. Moreover, define the cost of information as $ \mathrm{cost}_{info}(\hat{\mathbf{x}}_{t}):= \mathbf{e}^{T}_t\bar{\mathbf{W}}(\hat{\mathbf{x}}_{t})\mathbf{e}_t $ and cost of control effort as $ \mathrm{cost}_{eff}(\mathbf{u}_{t}):= \mathbf{u}_{t}^T\mathbf{V}_{t+1}^u\mathbf{u}_{t} $. Under the assumptions of holonomic systems and convex environment and given the initial re-sampled set of particles at time step $ t' $, $\{\mathbf{x}^{i}_{t'}\}^{k}_{i=1}$, and a goal state $ \mathbf{x}_g $, the core convex SLAP problem is:
\begin{align*}
\nonumber \min_{\mathbf{u}_{t':t'+K-1}}&\sum\limits_{t=t'+1}^{t'+K}[\mathrm{cost}_{info}(\hat{\mathbf{x}}_{t}) + \mathrm{cost}_{eff}(\mathbf{u}_{t-1})]\\
s.t.~~&\hat{\mathbf{x}}_{t'+K}=\mathbf{x}_g
\end{align*} 
where $ \hat{\mathbf{x}}_{t} = \tilde{\mathbf{A}}_{t':t-1}\hat{\mathbf{x}}_{t'}+\sum_{s=t'}^{t-1}\tilde{\mathbf{A}}_{s+1:t-1}(\mathbf{B}_s\mathbf{u}_{s}+f_s^p) $.
\end{problem}

\subsection{Static Environment with Non-Convex Constraints}\label{subsec:sol static environment}
In this subsection, we extend the solution of the previous subsection to include the non-convex constraints on the state, such as obstacles and banned areas in static environment with a known map of the environment. We define barrier functions to model the obstacles and softly incorporate them in the optimization objective. The optimization in such a case reduces to a locally convex optimization. Therefore, we need to initialize the optimization with a trajectory. This trajectory does not need to be feasible; however, starting from a feasible trajectory, the optimization avoids entering non-feasible states. Moreover, by initialing the optimization with trajectories in different homotopy classes, we find the locally optimal trajectories in different homotopy classes. We discuss the benefit of doing this towards the end of this subsection.

\textit{Polygonal obstacles approximated by ellipsoids:} Given a set of vertices that constitute a polygonal obstacle, we find the minimum volume enclosing ellipsoid (MVEE) and obtain its parameters \cite{moshtagh2005minimum}. Particularly, for the $ i $th obstacle, the barrier function includes a Gaussian-like function where the argument of the exponential is the MVEE, which can be disambiguated with its center $ \mathbf{c}^i $ and a positive definite matrix $ \mathbf{P}^i $ that determines the rotation and axes of the ellipsoid. Moreover, we add a line of infinity over the major and minor axes of the ellipsoid so the overall function acts as a barrier to prevent the trajectory from entering the region enclosed by the ellipsoid. Note that for non-polygonal obstacles, one can find the MVEE, and the algorithm works independently from this fact. Therefore, given the ellipsoid parameters $ \mathcal{C} := (\mathbf{c}^1, \mathbf{c}^2,\cdots, \mathbf{c}^{n_b})\in\mathbb{R}^{n_x\times n_b} $ and $ \mathcal{P} := (\mathbf{P}^1,\cdots, \mathbf{P}^{n_b})\in\mathbb{R}^{n^2_x\times n_b} $, the Obstacle Barrier Function (OBF) is constructed as follows:
\begin{align*} 
\nonumber \Phi^{(\mathcal{P},\mathcal{C})}(\mathbf{x})\!\!:&=M\!\sum_{i=1}^{n_b}[\exp(-[(\mathbf{x}-\mathbf{c}^i)^T\mathbf{P}^i(\mathbf{x}-\mathbf{c}^i)]^q)\\\nonumber
+\!\!\!\!\!\!\!\sum\limits_{\theta=0:\epsilon_m:1}\!\!\!\!\!\!(\!|\!|\mathbf{x}\!\!-\!(\theta \zeta^{i,1}\!&+\!(1\!\!-\!\theta) \zeta^{i,2})|\!|^{-\!2}_{2}\!\!+\!|\!|\mathbf{x}\!-\!(\theta \xi^{i,1}\!+\!(1\!\!-\!\theta) \xi^{i,2})|\!|^{-\!2}_{2})\!],
\end{align*}
where $ \epsilon_m = 1/m, m\in \mathbb{Z}^{+} $, $ M\in\mathbb{R}^{+} $, $ q\in\mathbb{Z}^{+} $, and $ \zeta^{i,1} $, $ \zeta^{i,2} $ and $ \xi^{i,1} $, $ \xi^{i,2} $ are the endpoints of the major and minor axes of the ellipsoid, respectively. Therefore, the second term in the sum places infinity points along the axes of the ellipsoid at points formed by convex combination of the two endpoints of each axis. As $ \epsilon_m $ tends to zero, the entire axes of the ellipsoid become infinite, and, therefore, act as a barrier to any continuous trajectory of states. One can think of putting more infinity points inside the ellipsoid by forming the convex combination of the existing infinity points. Moreover, the first term of the summand determines the territory of the ellipsoid and imposes an outwards gradient around the ellipsoid, acting as a penalty function pushing the trajectory out of the banned region. Hence, we define the cost of avoiding obstacles as:
\begin{align}\label{eq:cost obs integral}
{\mathrm{cost}}_{obst}(\mathbf{x}_{t1},\mathbf{x}_{t2}):=\int_{\mathbf{x}_{t1}}^{\mathbf{x}_{t2}}\Phi^{(\mathcal{P},\mathcal{C})}(\mathbf{x}')d\mathbf{x}',
\end{align}
which is the line integral of the OBF between two given points of the trajectory $ \mathbf{x}_{t1} $ and $ \mathbf{x}_{t2} $. Therefore, the addition of this cost to the optimization objective, ensures that the solver minimizes this cost and keeps the trajectory out of the banned regions. However, for implementation purposes, the integral in equation \eqref{eq:cost obs integral} is approximated by a finite sum consisting of fewer points between $ \mathbf{x}_{t1} $ and $ \mathbf{x}_{t2} $. 
Next, we provide the modified optimization problem that is locally convex in the inter-obstacle feasible space, and therefore can be solved using gradient descent methods \cite{boyd2004convex}.

\begin{problem}
\textbf{\textit{Locally convex problem in a static environment}}. Given $\{\mathbf{x}^{i}_{t'}\}^{k}_{i=1}$, $ \mathbf{x}_g $ and obstacle parameters $ (\mathcal{P},\mathcal{C}) $, the static environment problem for a holonomic system is:
\begin{align}\label{eq:main optim}
\nonumber \min_{\mathbf{u}_{t':t'\!+\!K\!-\!1}}&\!\!\sum\limits_{t=t'\!+\!1}^{t'\!+\!K}\!\!\![\mathrm{cost}_{info}(\hat{\mathbf{x}}_{t}\!)\!+\!\mathrm{cost}_{eff}(\!\mathbf{u}_{t-1}\!)\!+\! \mathrm{cost}_{obst}(\hat{\mathbf{x}}_{t-1},\hat{\mathbf{x}}_{t})]\\
s.t.~~&\hat{\mathbf{x}}_{t'+K}=\mathbf{x}_g.
\end{align}

\end{problem}
Moreover, we add convex saturation constraints of the type $ |\!|\mathbf{u}_t|\!| \le \max_{u} $ based on the specific robot model. 

Next, we proceed to optimize towards a better approximation among different homotopy classes while reaching predefined goals, such as uncertainty reduction, collision avoidance, and reaching the final destination with minimal energy effort.

\begin{figure}[ht!]
  \centering
  {\includegraphics[width=1.3in]{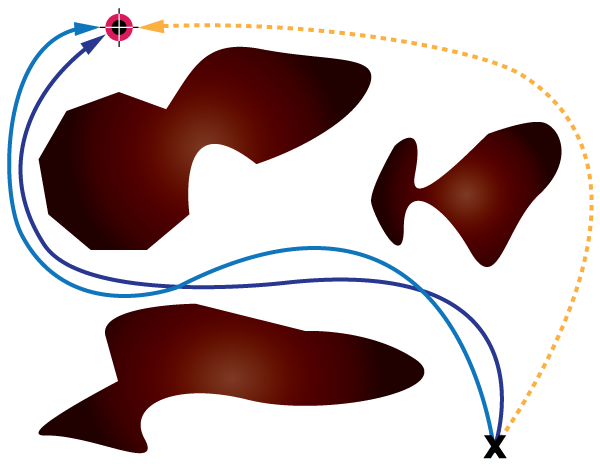}}
  \caption{Homotopy classes. The solid trajectories are in a different homotopy class from the dashed trajectory.\label{fig:homotopy} }\vspace{-14pt}
\end{figure}

\textit{Homotopy classes:} Homotopy classes of trajectories are defined as sets of trajectories that can be transformed into each other by a continuous function without colliding with obstacles \cite{bhattacharya2010search,bhattacharya2012topological}. As shown in Fig. \ref{fig:homotopy} the two solid trajectories are in one homotopy class, while the dashed trajectory is in a different class.

\textit{Homotopy classes and optimal trajectory:} There are several methods to find the trajectories in homotopy classes \cite{bhattacharya2010search,bhattacharya2012topological}. For instance, in low dimensions one can construct the visibility graph considering the pure motion planning problem and find trajectories in different homotopy classes that connects the start state to the goal state by pruning the non-unique paths. These methods provide such paths for different purposes such as finding the shortest path. However, usually the uncertainty or dynamics of the system are not considered. We initialize our optimization with non-looped trajectories in different homotopy classes \cite{bhattacharya2012topological}. The optimizer considers the cost of uncertainty, effort, and collision-avoidance along with the linearized dynamics of the (holonomic) system and morphs the initial trajectory towards a locally optimal trajectory. Our barrier function model of the obstacles prevents the trajectory from entering the banned regions. These barrier functions, along with a optimization tuned through the saturation constraints, a long enough optimization horizon (determined by the time-discretization step of the initial trajectory), and a limited step size of the line-search in optimization \cite{yakowitz2012algorithms,van2012motion}, keep the trajectory in its initial homotopy class. Moreover, since the optimization is locally convex, it finds the local optimal trajectory of that homotopy class under the imposed constraints and conditions starting from a trajectory in that class. Therefore, by comparing the total costs obtained in different cases, we obtain the lowest cost smooth trajectory considering all the predefined costs, and most significant of all, uncertainty reduction. This is the closest output trajectory of our algorithm to the globally optimal trajectory in the existence of uncertainties.
\subsection{Dynamic Environment with Time-Varying Constraints}\label{sec:sol dynamic environment}

Now that we have specified all the machinery needed to find the optimal path in terms of the defined cost in a static environment, we extend our method to an environment that is not fully static. 

\textit{Incorporating dynamic obstacles:} If some of the obstacles are moving, the state constraints become time-varying. In such a case, we modify the optimization problem by altering the obstacle cost so it includes the dynamic obstacles as follows:
\begin{align*} 
\nonumber \!\!\!\Phi^{(\mathcal{\hat{P}}_t,\mathcal{\hat{C}}_t)}_{t}(\mathbf{x})\!\!:=&M\!\sum_{i=1}^{n_b}[\exp(-[(\mathbf{x}-\mathbf{\hat{c}}^i_{t})^T\mathbf{\hat{P}}^i_{t}(\mathbf{x}-\mathbf{\hat{c}}^i_{t})]^p)\\\nonumber
+\!\!\!\!\!\!\!\sum\limits_{\theta=0:\epsilon_m:1}\!\!\!\!\!\!(\!|\!|\mathbf{x}\!\!-\!(\theta \hat{\zeta}^{i,1}_t\!&+\!(1\!\!-\!\theta) \hat{\zeta}^{i,2}_t)|\!|^{-\!2}_{2}\!\!+\!|\!|\mathbf{x}\!-\!(\theta \hat{\xi}^{i,1}_t\!+\!(1\!\!-\!\theta) \hat{\xi}^{i,2}_t)|\!|^{-\!2}_{2})\!],
\end{align*}
where $ \mathbf{\hat{c}}^i_{t} $, $ \mathbf{\hat{P}}^i_{t} $, $ \hat{\zeta}^{i,1}_{t} $, $ \hat{\zeta}^{i,2}_{t} $, $ \hat{\xi}^{i,1}_{t} $ and $ \hat{\xi}^{i,2}_{t} $ are the estimated parameters of the $ i $th obstacle at time step $ t $ given by a separate estimator that tracks the obstacles. Note that if the $ i $th obstacle is just moving, then at time $ t'>t $, $ \mathbf{\hat{c}}^i_{t'}= \mathbf{\hat{c}}^i_{t} +\mathbf{\hat{v}}^i(t'-t) $ and $ \mathbf{\hat{P}}^i_{t'} = R^i_{\hat{\alpha}}\mathbf{\hat{P}}^i_{t} $ where $ \mathbf{\hat{v}}^i $ is a constant estimated velocity vector, and $ R^i_{\hat{\alpha}} $ is an estimated rotation matrix by $ \hat{\alpha} $ degrees. However, if there is a change of shape in the obstacle or appearance of new obstacles, we run the MVEE algorithm to find the parameters of that obstacle. For our planning purposes, we assume there is a separate estimator that tracks and estimates the obstacles' parameters, and our planner only uses the results obtained by that tracker to find the optimized trajectory. Moreover, since the algorithm is implemented in an RHC fashion, if there is a change in the estimates of the obstacles, for the next step the optimization uses the new estimates of the obstacle parameters. Moreover, the obstacle cost is modified as follows:
\begin{align*}
\mathrm{cost}_{obst}(\mathbf{x}_{t1},\mathbf{x}_{t2}, t)\!\!:=\int_{\mathbf{x}_{t1}}^{\mathbf{x}_{t2}}\Phi^{(\mathcal{\hat{P}}_{t},\mathcal{\hat{C}}_{t})}_{t}(\mathbf{x}')d\mathbf{x}'.
\end{align*}\label{eq:approximate obstacle cost dynamic}

\begin{problem}
\textbf{\textit{Dynamic environment}}. For a holonomic system, given $\{\mathbf{x}^{i}_{t'}\}^{k}_{i=1}$, $ \mathbf{x}_g $ and estimates of the obstacle parameters for the entire lookahead horizon $ \{(\mathcal{\hat{P}}_t,\mathcal{\hat{C}}_t)\}_{t=t'}^{t'+K+1} $, the dynamic environment problem is defined as:
\begin{align}\label{eq:dynamic environment}
\nonumber \min_{\mathbf{u}_{t':t'+K-1}}&\sum\limits_{t=t'+1}^{t'+K}[\mathrm{cost}_{info}(\hat{\mathbf{x}}_{t}) + \mathrm{cost}_{eff}(\mathbf{u}_{t-1})\\\nonumber&~~~~~~~~+ \mathrm{cost}_{obst}(\hat{\mathbf{x}}_{t-1},\hat{\mathbf{x}}_{t}, t-1)]\\
s.t.~~&\hat{\mathbf{x}}_{t'+K}=\mathbf{x}_g.
\end{align}
\end{problem}
If there is a sudden appearance of a new obstacle in part of the trajectory, only that part of the trajectory is changed provided there is still a feasible path between the two points immediately outside and on the other side of that obstacle. Otherwise, the entire algorithm runs again from the current state to the goal state. It should be added that, unlike a static environment, in a stochastic problem with dynamic environment, unless the planning horizon is very small, there is not much that can be said regarding the homotopy paths discussed in Section \ref{subsec:sol static environment}. This is an ongoing research. 

Now that we have provided our solution for all the three cases, we proceed to discuss the implementation strategy.

\subsection{RHC Implementation}\label{subsec:RHC Implementation}

\begin{figure}[!t]
\centering
   \includegraphics[width=2.7in]{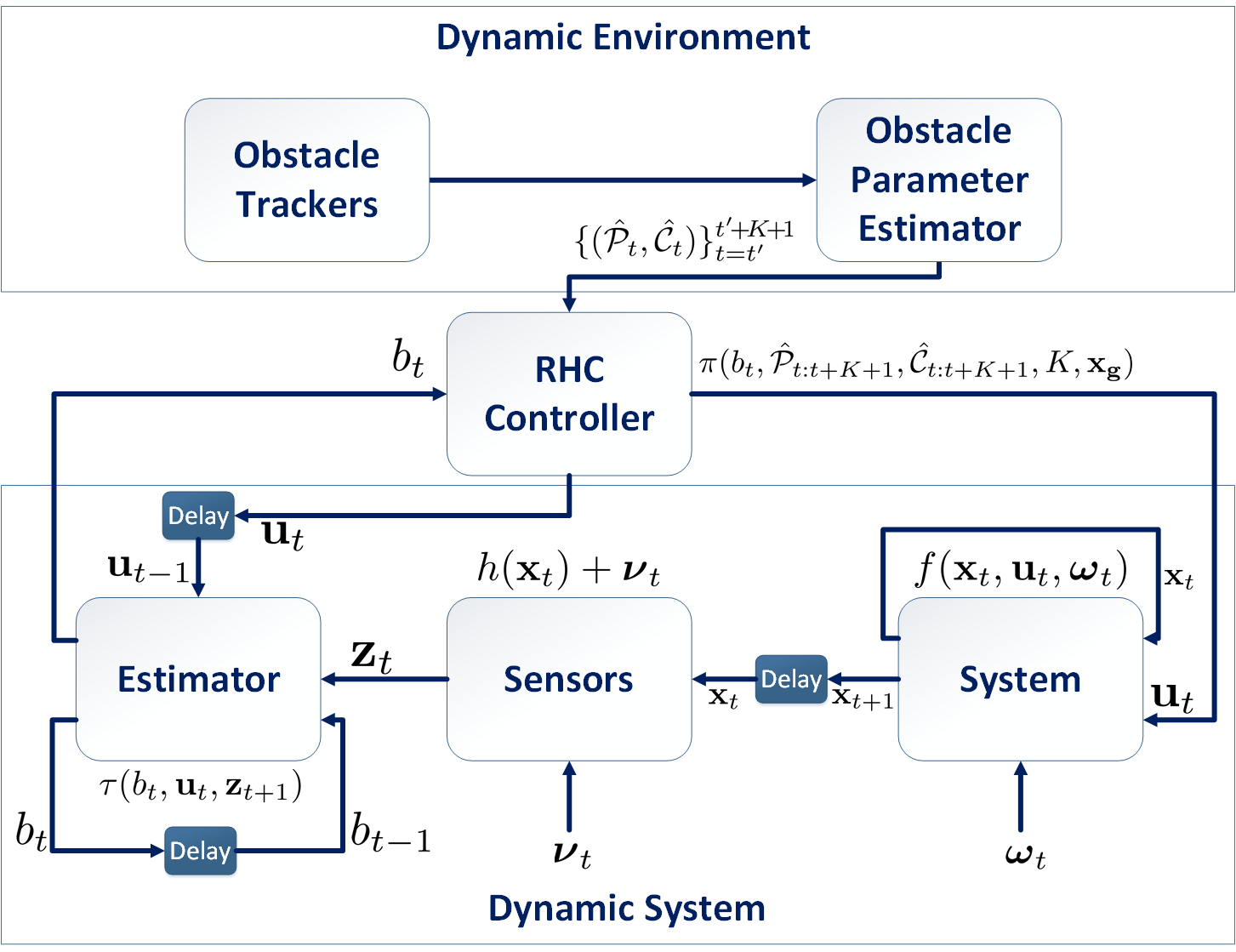}

  \caption{The overall feedback control loop.}
  \label{fig:RHC_Loop}\vspace{-20pt}
\end{figure}

The overall feedback control loop is shown in Fig. \ref{fig:RHC_Loop}. The system initiates from a non-Gaussian distribution in the feasible state space that constitutes the initial belief. In the case of a dynamic environment, the most complicated case of our problems, given the current belief, $ b_t $, estimates of the obstacles' parameters, $ \{(\mathcal{\hat{P}}_t,\mathcal{\hat{C}}_t)\}_{t=t'}^{t'\!+\!K\!+\!1} $, lookahead time horizon, $ K $, and the goal state, $ \mathbf{x}_g $, the RHC policy function $ \pi:\mathbb{B}\times\mathbb{R}^{n_x\times n_b\times (K\!+\!1)}\times\mathbb{R}^{n^2_x\times n_b\times (K\!+\!1)}\times\mathbb{R}\times\mathbb{X}\rightarrow\mathbb{U} $ generates an optimal action $ \mathbf{u}_{t}=\pi(b_t, \mathcal{\hat{P}}_{t:t+K+1},\mathcal{\hat{C}}_{t:t+K+1}, K, \mathbf{x_g}) $, which is the first element of the open-loop optimal sequence of actions generated in different variants of problem \ref{problem:General optimal control problem}. The agent executes $ \mathbf{u}_{t} $ transitioning the state of the system from $ \mathbf{x}_{t} $ in $ \mathbf{x}_{t+1} $ where a new observation $ \mathbf{z}_{t+1} $ is obtained by the sensors. The estimator updates the belief as $ b_{t+1}\!\!\!=\!\!\!\tau(b_{t},\mathbf{u}_t,\mathbf{z}_{t+1}) $ and the policy is fed the updated belief to close the loop. Meanwhile, on another separate loop, the obstacle trackers measure the current state of the obstacles and the obstacle parameter estimators obtain the estimates of the obstacles. As mentioned above, the estimates are fed into the policy function immediately before the controller plans its next action. In the case of the static environment, the policy function is fed the parameters of the obstacles that remain the same for the entire horizon. Similarly, in the case of a convex environment, the general boundaries and convex constraints take the place of the obstacle parameters in the planning problem.

\textit{Stopping execution:} The algorithm stops when the probability of reaching the goal, calculated as the area under the belief density over the goal region, exceeds a predefined value \cite{platt2013convex}.

The planning algorithm is in Algorithm \ref{alg:re-planning}.\vspace{-8pt}
\begin{algorithm}
    \SetKwInOut{Input}{Input}
    \Input{Initial belief state $b_{t'}$, Goal state $\mathbf{x}_{g}$, Planning horizon $K$, Belief dynamics $\tau$, Obstacle parameters $ \{(\mathcal{\hat{P}}_t,\mathcal{\hat{C}}_t)\}_{t=t'}^{t'+K+1} $}
    \While{$\mathcal{P}(b_t,r,\mathbf{x}_g)\le \breve{w}_{th}$}{
    $ \mathbf{u}_{t}\gets\pi(b_t, \mathcal{\hat{P}}_{t:t+K+1},\mathcal{\hat{C}}_{t:t+K+1}, K, \mathbf{x_g}) $\;
    execute $ \mathbf{u}_{t} $, perceive $ \mathbf{z}_{t} $\;
    $b_{t+1}(\mathbf{x})\gets\tau(b_{t}(\mathbf{x}),\mathbf{u}_{t},\mathbf{z}_{t+1})$\;
    }
\caption{Planning Algorithm}\label{alg:re-planning}
\end{algorithm}\vspace{-18pt}

\subsection{A Discussion and Comparison on Complexity}\label{subsec:Discussion on Complexity}
The core optimization problem in a convex environment as defined in problem \ref{problem:Core convex problem in convex feasible space} is a convex program that does not require an initial solution at all. The number of decision variables is $ Kn_u $, and there is $ 1 $ linear equality constraint, plus, the robot's saturation inequality constraints, which can be $ Kn_u $ at most. Therefore, the optimization involves the minimum number of decision variables. Let us assume for simplicity that the size $ \mathbf{x} $,  $\mathbf{u} $, and $ \mathbf{z} $ vectors are all $ O(n) $. Thus, utilizing a common method, such as center of gravity for convex optimization \cite{bubeck2014theory} to obtain a \textit{globally optimal} solution with $ \epsilon $ confidence, the algorithm requires $ \Omega(Knlog(1/\epsilon)) $ calls to the oracle \cite{nemirovsky1983problem}. On the other hand, in equation \eqref{eq:for complexity}, $ \tilde{\mathbf{A}}_{t':t-1}^T\mathbf{W}(\hat{\mathbf{x}}_{t})\tilde{\mathbf{A}}_{t':t-1} $ requires a multiplication of $ O(n)\times O(n) $ matrices for $ O(K) $ times, which takes $ O(Kn^3) $. However, the multiplication of the vectors $ (\mathbf{x}^i_{t'}-\hat{\mathbf{x}}_{t'}) $ to a $ \mathbb{R}^{n\times n} $ matrix involves $ O(Nn^2) $ time. The outer sum also takes $ K $ time. All the other operations, such as calculation of $ \hat{\mathbf{x}}_t $ and constraints, take less time. Hence, the time complexity of the computations is $ O(Kn^3+Nn^2) $. In LQG-based belief space methods that construct a trajectory, the method described in \cite{Berg11-isrr} involves a non-convex optimization, which takes $ O(K n^6) $ computations with a second-order convergence rate to a locally optimal solution. Another RHC-based method particle filter-based method is described in \cite{platt2013convex}, where the core problem is a convex problem in $ Kn_u+N $ number of decision variables with $ N(K-1)+1 $ number of inequality constraints. The algorithm assumes a linear process model with Gaussian noise and a linear measurement model with a Gaussian noise whose covariance is state-dependent. The solution is categorized in the second class of methods in Fig. \ref{fig:belief_propagarion}. Moreover, to include more than one observation source, the algorithm requires a modification with integer programming, such that at each time step, there could only be one observation. Although the analysis of time complexity is not given, to the best of our knowledge we calculate it to be $ O(NK(Kn^3+Nn^2)) $ without integer programming. Moreover, the convergence needs $ \Omega((N+Kn)log(1/\epsilon)) $ calls. 
In the presence of obstacles, the size of our optimization does not change; however, the rate of convergence reduces to the rate of gradient descent methods. Furthermore, the solution becomes locally optimal starting with an infeasible solution whose immediate gradient is not towards the local minima of the obstacles. Theoretically, if the $ \epsilon_m $ of OBF tends to infinity, there is no local minima of the barriers; nevertheless, practically, starting from a semi-feasible trajectory, a tuned optimization results in convergence to a locally optimal feasible solution. In \cite{Berg11-isrr}, in the presence of obstacles, the convergence rate and computational cost does not change, but the (tuned) optimization must start with a feasible path. In \cite{platt2013convex}, obstacles are modeled with a chance-constrained method that involves the introduction of additional variables and integer programming with iterative applications of the algorithm. This limits the scalability of that method to complex environments. 
\vspace{-10pt}
\section{Simulations and Examples}\label{sec:simulations}
\begin{figure}%
\centering
\subfloat[]{\includegraphics[width=1.5in]{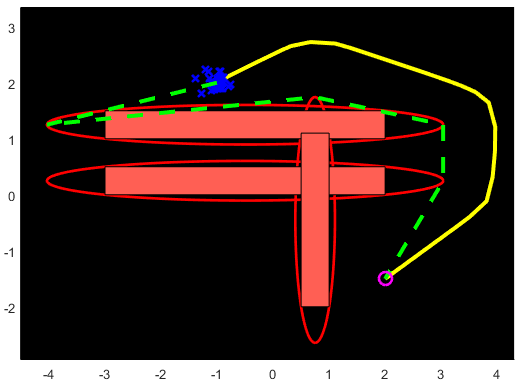}}\qquad
\subfloat[]{\includegraphics[width=1.5in]{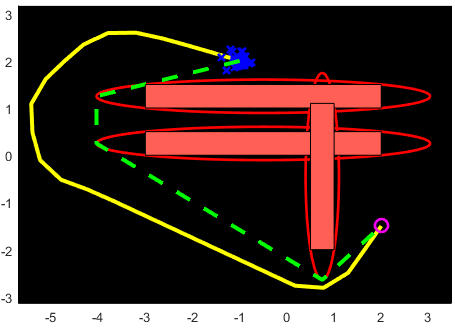}}\\
\subfloat[]{\includegraphics[width=1.5in]{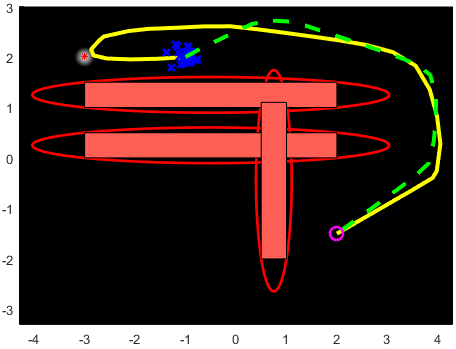}}\qquad
\subfloat[]{\label{4figs}\includegraphics[width=1.45in]{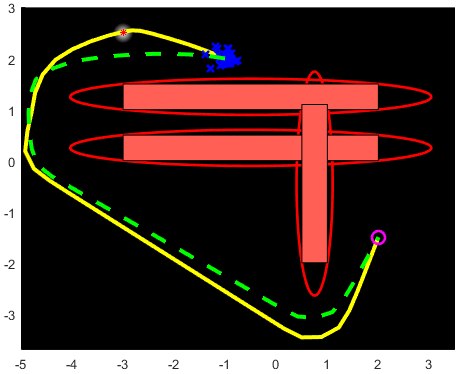}}%
\caption{Four cases. Cases a and b show the resulting paths generated by optimizing without considering the information sources, whereas cases c and d consider information sources.}
\label{3figs}\vspace{-18pt}
\end{figure}

In this section, we exhibit some applications for our method. We performed our simulations in MATLAB 2015b in a 2.90 GHz CORE i7 machine with dual core technology and 8 GB of RAM. First, we perform a comparative simulation in which we compare the results of our algorithm for a static environment with and without information sources. Then, we perform an experiment for a KUKA youBot in static environment. Finally, we perform simulations for dynamic environments. There are 8 multimedia files uploaded for figures \ref{3figs}, \ref{fig:YouBot base control}, \ref{fig:dyn}, \ref{fig:dyn elliptic} that include more details on the simulations. In these figures, all the axes units are in meters.

\begin{figure}[ht!]
  \centering
  {\includegraphics[width=1.6in]{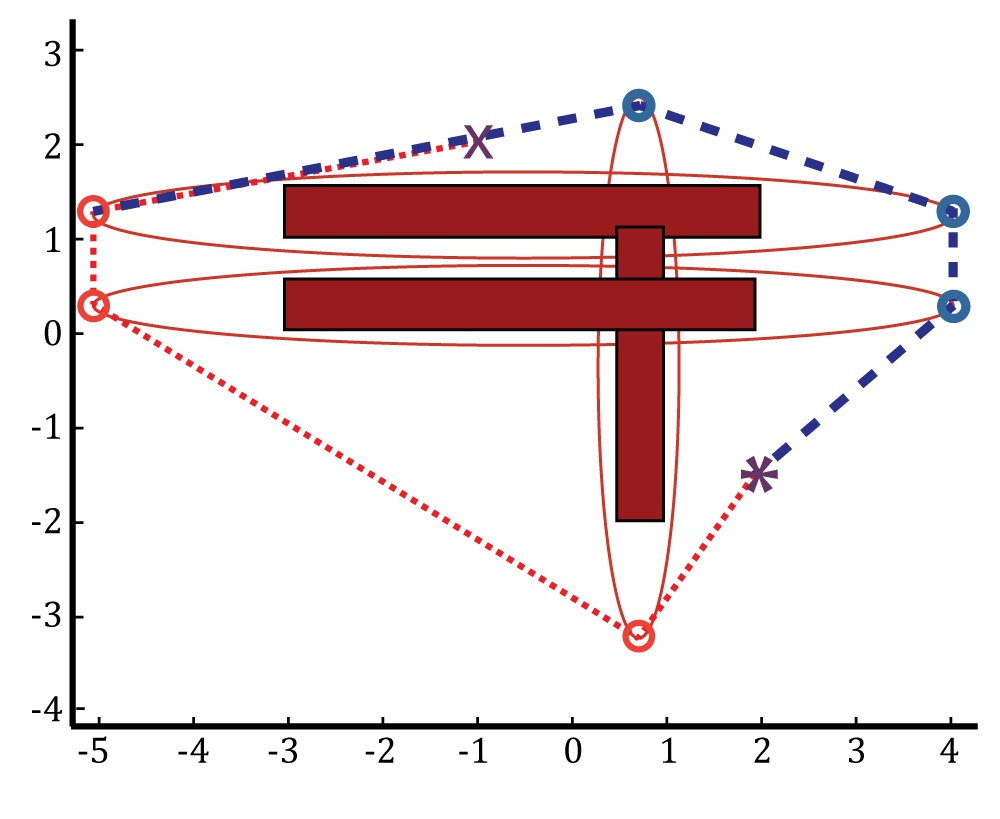}}
  \caption{Modified visibility graph. The two non-looped non-homotopic trajectories are found with modified visibility graph and are indicated by the red dotted and blue dashed paths.\label{fig:visiblity} }\vspace{-14pt}
\end{figure}
\subsection{Simple Comparison Test}
Figure \ref{fig:visiblity} shows an environment with three obstacles forming a connected obstacle. The banned areas are enclosed with three MVEEs. In this experiment, we use the visibility graph to find initial trajectories in different homotopy classes. Moreover, instead of using the polygons, we use the ellipsoids that enclose them. Since our optimization utilizes a gradient descent method, we only consider the straight lines between the nodes and ignore the collision of the straight line with the ellipsoid that the node is lying on. This increases the speed of finding the visibility graph and coupled with optimization over the output paths, the minor collisions do not hurt the algorithm. 

Next, each of the two paths is discretized to satisfy the tuning properties described in Section \ref{subsec:sol static environment}. They are then fed into the optimization function $ \pi $ to produce the optimized smooth collision-free paths. We have produced two sets of results; in the first set, we do not consider the cost of information (as if we are considering the motion planning problem to generate smooth collision-free paths); in the second, we add a landmark as the information source, and consider the optimization with cost of information, to compare the results. As seen in Fig. \ref{4figs}, existence of the landmarks changes the paths of the robot, such that the robot visits them to reduce its uncertainty and then continues its path towards the goal state. 

\subsection{KUKA YouBot}
\begin{figure}\vspace{-10pt}
\centering
{\includegraphics[width=2in]{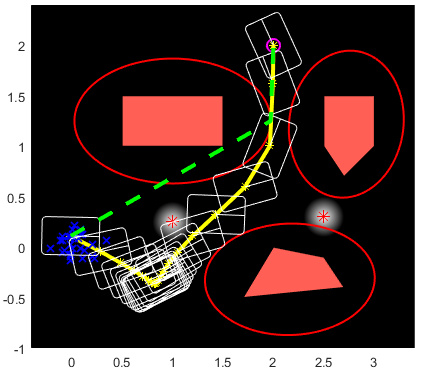}}
\caption{YouBot base control. There are three obstacles and two landmarks. The robot base is shown by a rectangle with a line at the heading. Initial and planned trajectories are depicted by dashed and solid lines, respectively.}\vspace{-10pt}
\label{fig:YouBot base control}
\end{figure}

In this section, we use the kinematics equations of KUKA youBot base as described in \cite{1_zakharov_2011}. Particularly, the state vector can be described by a 3D vector, $ \mathbf{x} = [\mathbf{x}_{\mathtt{x}}, \mathbf{x}_{\mathtt{y}}, \mathbf{x}_{\theta} ]^T $, describing the position and heading of the robot base, and $ \mathbf{x}\in SO(3) $. The control consists of the velocities of the four wheels. It can be shown that the discrete motion model can be written as $ \mathbf{x}_{t+1} = f(\mathbf{x}_{t}, \mathbf{u}_t, \boldsymbol{\omega}_t) = \mathbf{x}_t+\mathbf{B}\mathbf{u}_tdt+\mathbf{G}\boldsymbol{\omega}_t\sqrt{dt} $, where $ \mathbf{B} $ and $ \mathbf{G} $ are appropriate constant matrices whose elements depend on the dimensions of the robot as indicated in \cite{2_youbot_store_com_2016}, and $ dt $ is the time-discretization period. Inspired by \cite{zucker2013chomp}, we model the robot with a configuration of a set of points which represent the centers of the balls that cover the body of the robot. In our method, we cover the robot with two balls whose radii are proportional to the width of the robot. We find the MVEE of the polygons  that are inflated from each vertex to the size of the radius and modify the cost of obstacles to keep the centers of the balls out of the new barriers. The observation model is a range and bearing based model where the corresponding elements of the $ \mathbf{R} $ matrix are chosen to be $ |\!|(\mathbf{x}_{\mathtt{x}}-L_x, \mathbf{x}_{\mathtt{y}}-L_y)|\!|_2^2 $ for all observations so as to have the desired features described in Lemma \ref{lemma 1}. $ (L_x, L_y) $ represents the coordinates of a landmark. The results depicted in Fig. \ref{fig:YouBot base control}, show that the planned trajectory avoids entering the banned regions bordered by the ellipsoids, so that the robot itself avoids colliding with the three obstacles.

\subsection{Dynamic Environment}
\begin{figure}[ht!]\vspace{-10pt}
  \centering
  \subfloat[]{\includegraphics[width=1.5in]{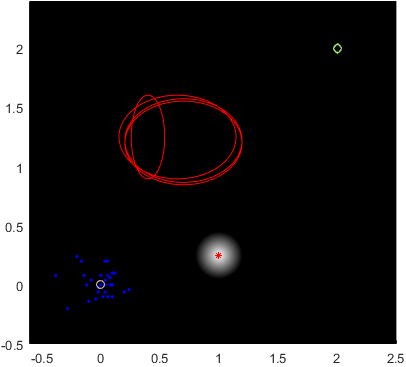}}\qquad
      \subfloat[]{\includegraphics[width=1.45in]{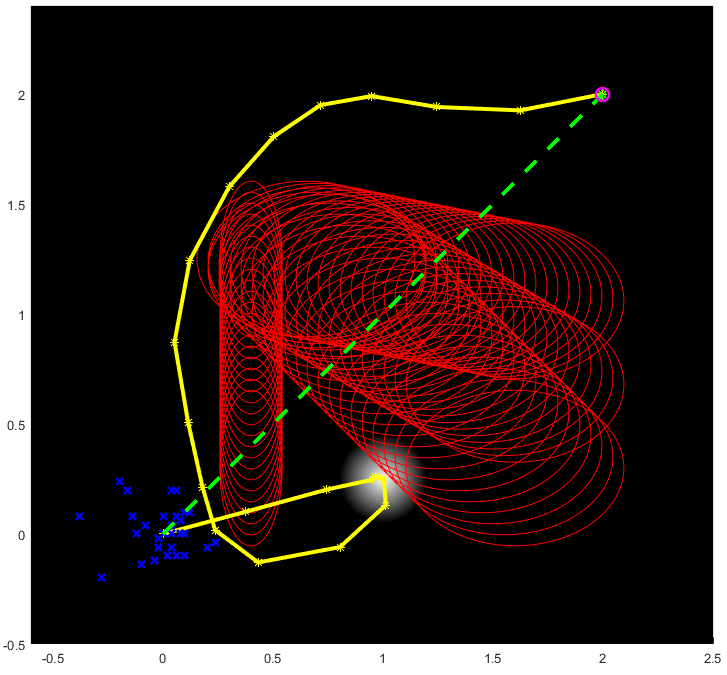}}
  \caption{Dynamic Environment. The robot heads towards the landmark to reduce its uncertainty and avoids the moving objects by changing its path to be in the opposite direction of the objects.\label{fig:dyn} }\vspace{-16pt}
\end{figure}

In this scenario, we simulate a case where there are four objects, starting from a common position and moving in different directions downwards and towards the right of the map. The robots starts from a distribution whose mean is at (0,0), and wishes to reach the goal state (2,2) with high probability. As seen in Fig. \ref{fig:dyn}, at the beginning of its trajectory, the robot head towards the landmark at (1, 0.5), and as the moving obstacles get closer, it changes its direction to bypass the objects in the opposite direction. In this scenario, the initial trajectory is just the straight line between the most probable initial location of the robot and the final destination shown in the figure with green dashed line, with the planned trajectory of the robot shown as a solid yellow line.

\begin{figure}[ht!]\vspace{-8pt}
  \centering
  \subfloat[]{\includegraphics[width=1.5in]{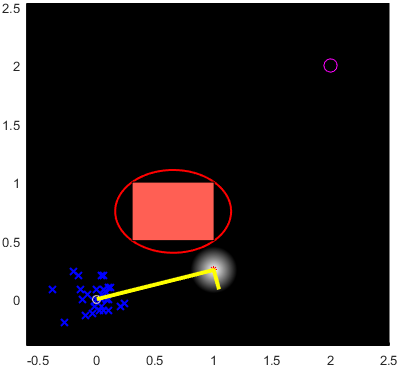}}\qquad
      \subfloat[]{\includegraphics[width=1.45in\label{subfig:elliptic_final}]{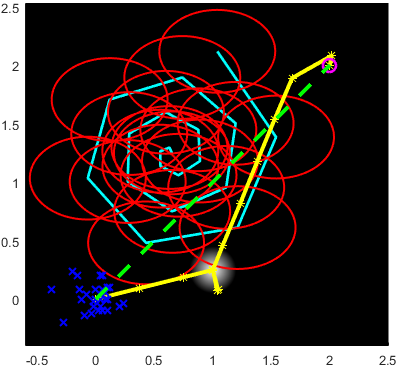}}
  \caption{Moving object. The robot spends most of its time near the information source and avoids the object, which is moving in a spiral path, and heads towards the goal region safely.\label{fig:dyn elliptic} }\vspace{-12pt}
\end{figure}

In another scenario shown in Fig. \ref{fig:dyn elliptic}, an object is moving in a spiral path shown in Fig. \ref{subfig:elliptic_final} with the robot trying to avoid colliding with the obstacle, spending most of its time near the information source and reaching the goal in a safe, short and smooth path.
\section{Conclusion}\label{sec:conclusion}
In this paper, we have proposed a method for controlling a stochastic system starting from a configuration in the state space to reach a goal region. We have proposed a barrier function method that combined with our optimization and gradient descent methods, such as an interior point method, keeps a trajectory in its homotopic path, and finds the local optimal path. The optimization seeks a smooth trajectory that results in the lowest combined cost of uncertainty in robot's perception and energy effort, while avoiding collision with obstacles, and reaching to the goal destination with high probability. Moreover, by finding trajectories in different homotopy classes and optimizing over paths in all classes, we can compare the costs incurred by those trajectories. Therefore, we find the lowest cost trajectory that considers the cost of uncertainty incurred in the trajectories, which is the closest trajectory to the global optimal trajectory while avoiding the need to solve the belief space dynamic programming equations. Finally, our receding horizon control strategy and the low computational cost of the optimization allows us to incorporate dynamic obstacles, as well as new objects that are detected along the trajectory of the robot by adjusting the optimization and the barrier functions during execution. This allows us to perform belief space planning in an online manner. Our future work is aimed at extending our results, analyzing the performance and sensitivities, and executing the methods on real robots.

\section*{Acknowledgment}
\addcontentsline{toc}{section}{Acknowledgment}
This material is based upon work partially supported by NSF under Contract Nos. CNS-1646449 and Science \& Technology Center Grant CCF-0939370, the U.S. Army Research Office under Contract No. W911NF-15-1-0279, and NPRP grant NPRP 8-1531-2-651 from the Qatar National Research Fund, a member of Qatar Foundation.

\bibliographystyle{plainnat}
\bibliography{AliAgha}

\end{document}